\documentclass[sigconf,screen]{acmart}
\usepackage{amsmath}
\usepackage{indentfirst}
\usepackage{enumitem}
\usepackage{fancyhdr}
\usepackage{hyperref}
\usepackage{multirow}
\usepackage{multicol}
\usepackage{appendix}
\usepackage{amssymb}
\newcommand{\etal}{\textit{et al.}}
\newcommand{\eg}{\emph{e.g.}}
\newcommand{\ie}{\emph{i.e.}}
\newcommand{\etc}{\emph{etc}}

\AtBeginDocument{%
  \providecommand\BibTeX{{%
    \normalfont B\kern-0.5em{\scshape i\kern-0.25em b}\kern-0.8em\TeX}}}

\copyrightyear{2020}
\acmYear{2020}
\setcopyright{acmcopyright}\acmConference[MM '20]{Proceedings of the 28th ACM International Conference on Multimedia}{October 12--16, 2020}{Seattle, WA, USA}
\acmBooktitle{Proceedings of the 28th ACM International Conference on Multimedia (MM '20), October 12--16, 2020, Seattle, WA, USA}
\acmPrice{15.00}
\acmDOI{10.1145/3394171.3413981}
\acmISBN{978-1-4503-7988-5/20/10}

\begin{document}

\fancyhead{}

\title{Dual In-painting Model for Unsupervised Gaze Correction and Animation in the Wild}


\author{Jichao Zhang$^1$,\, Jingjing Chen$^2$,\, Hao Tang$^1$,\, Wei Wang$^3$,\, Yan Yan$^4$,\, Enver Sangineto$^1$,\, Nicu Sebe$^{1,5}$}
\affiliation{%
 \institution{$^1$University of Trento, $^2$Shandong University, $^3$\'Ecole Polytechnique F\'ed\'erale de Lausanne \\
  $^4$Texas State University, $^5$Huawei Research Ireland}
}



\begin{abstract}

In this paper we address the problem of unsupervised gaze correction in the wild, presenting a solution that works without the need for precise annotations of the gaze angle and the head pose. We have created a new dataset called CelebAGaze, which consists of two domains $X$, $Y$, where the eyes are either staring at the camera or somewhere else. Our method consists of three novel modules: the Gaze Correction module~(GCM), the Gaze Animation module~(GAM), and the Pretrained Autoencoder module~(PAM). Specifically, GCM and GAM separately train a dual in-painting network using data from the domain $X$ for gaze correction and data from the domain $Y$ for gaze animation. Additionally, a Synthesis-As-Training method is proposed when training GAM to encourage the features encoded from the eye region to be correlated with the angle information, resulting in a gaze animation which can be achieved by interpolation in the latent space.
To further preserve the identity information~(\eg, eye shape, iris color), we propose the PAM with an Autoencoder, which is based on  Self-Supervised mirror learning where the bottleneck features are angle-invariant and which works as an extra input to the dual in-painting models. Extensive experiments validate the effectiveness of the proposed method for gaze correction and gaze animation in the wild and demonstrate the superiority of our approach in producing more compelling results than state-of-the-art baselines. Our code, the pretrained models and the supplementary material are available at: \url{https://github.com/zhangqianhui/GazeAnimation}.

\end{abstract}



\maketitle

\section{Introduction}
Gaze correction aims at manipulating the eye gaze with respect to the desired direction. This is important in some real-life scenarios where there is the need to  ``move'' the person's gaze into the camera. For example, shooting a good portrait is challenging as the subject may be too nervous to stare at the camera. Another scenario is videoconferencing where the eye contact is extremely important as the gaze can express attributes such as attentiveness and confidence. Unfortunately, eye contact and gaze awareness are lost in most videoconferencing systems, as the participants look at the monitors and not directly into the camera.

\begin{figure}[t]
\begin{center}
\includegraphics[width=0.9\linewidth]{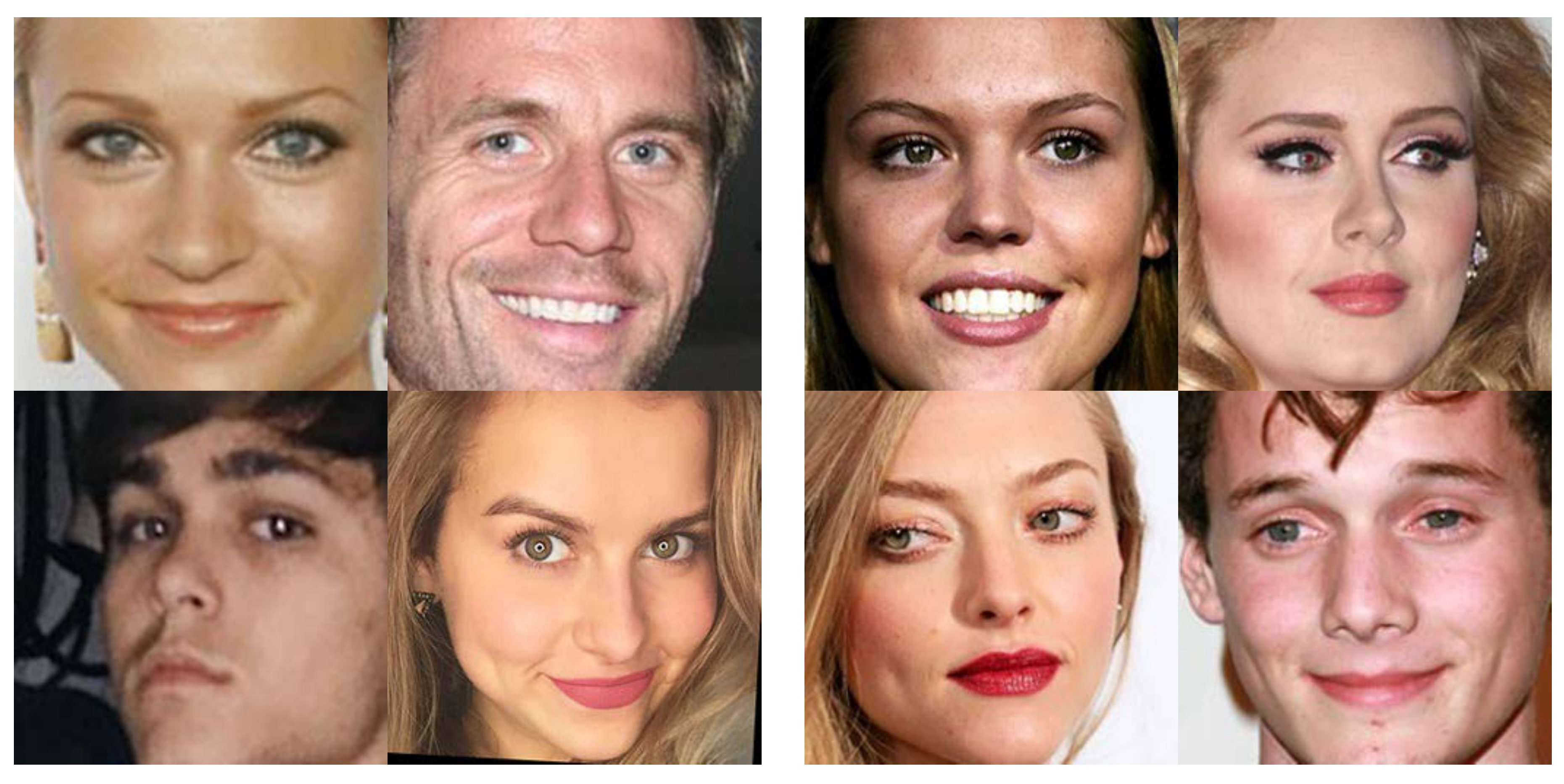}
\end{center}
\vspace{-0.5cm}
\caption{A few example images of the CelebAGaze dataset. Left: People staring at the camera for domain $X$. Right: People staring somewhere else for domain $Y$. These examples show the diversity in the head pose, the gaze angle and the illumination conditions of the proposed dataset.}
\label{fig:newgaze}
\vspace{-0.5cm}
\end{figure}

Early works in gaze correction relied on special hardware, such as stereo cameras~\cite{criminisi2003gaze,yang2002eye}, Kinect sensor~\cite{kuster2012gaze} or transparent mirrors~\cite{kollarits199634,okada1994multiparty}. Recently, some learning-based methods have produced very high-quality synthetic images with a corrected gaze. For instance, Kononenko and Lempitsky~\cite{kononenko2015learning} propose to solve the problem of monocular gaze correction using decision forests. DeepWarp~\cite{ganin2016deepwarp} uses a deep network to directly predict an image-warping flow field with a coarse-to-fine learning process. However, despite its good qualitative results, this method  fails in generating photo-realistic images when redirecting the gaze for large angles. Additionally, it produces unnatural eye shapes as the $L1$ loss is exploited to learn the flow field without any geometry regularization. To solve this problem, PRGAN~\cite{he2019photo} proposes to exploit adversarial learning with a cycle-consistent loss to generate more plausible gaze redirection results. However, it is challenging for these methods~\cite{kononenko2015learning,ganin2016deepwarp,he2019photo} to obtain high-quality gaze redirection results in the wild when there are large variations in the head pose, gaze angles \etc. Another category of works is based on a 3D model without training data such as GazeDirector~\cite{wood2018gazedirector}. The main idea of GazeDirector is to model the eye region in a 3D reference system instead of predicting a flow field directly from the input image. However, modeling in 3D has strong assumptions that do not hold in reality.

More importantly, the previous methods need training labels of the head pose and gaze angles, but these are hard to obtain in the wild. To solve these problems, we collected the CelebAGaze dataset which consists of two domains: eyes staring at the camera for domain $X$ and eyes staring somewhere else for domain $Y$, as shown in Fig.~\ref{fig:newgaze}. Note that no paired samples exist in the CelebAGaze dataset. We propose an unsupervised learning method for gaze correction and animation consisting of three modules: 1) GCM is an in-painting model, trained on the domain $X$, which learns how to fill in the missing eye regions with the new content which has the corrected eye gaze; 2) GAM exploits the other in-painting model, trained on the domain $Y$. To generalize the gaze animation with various angle directions, we propose the Synthesis-As-Training method to use synthesizing data for training GAM, encouraging the features encoded from the eye region to be corrected with the angle information. Finally, gaze animation can be achieved by interpolating this feature in the latent space; 3) PAM encodes the angle-invariant content features, \eg, iris color, eye shape. Specifically, we pretrain an autoencoder by self-supervised mirror learning where the bottleneck features are extracted as an extra input of the dual in-painting model to preserve the identity of the corrected results. Finally, both the qualitative and quantitative evaluations demonstrate that our model achieves more compelling results than the state-of-the-art baselines in gaze correction and gaze animation. To recap, our main contributions are:
\begin{itemize}[leftmargin=*]
    \item[1)] We propose a simple yet effective dual in-painting method for gaze correction and animation.
    \item[2)] For gaze animation, we introduce a novel Synthesis-As-Training Method which encourages the features encoded from the eye region to be correlated with the angle information.
    \item[3)] We design a novel Self-Supervised Mirror Learning for pretraining an autoencoder to extract content features to preserve the identity of the corrected results.
    \item[4)] We make available to the research community a new gaze dataset for gaze correction and animation.
\end{itemize}

\begin{figure*}[ht]
\begin{center}
\includegraphics[width=0.9\linewidth]{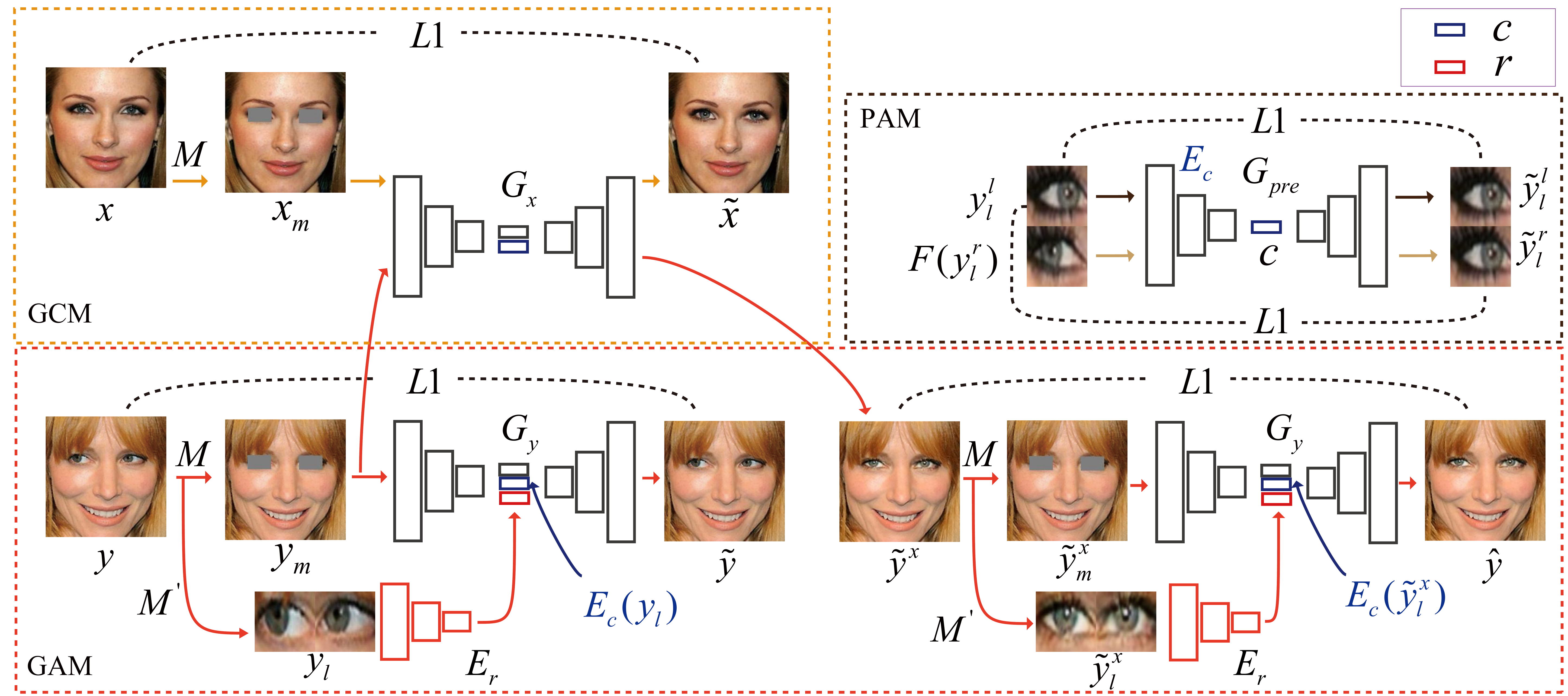}
\end{center}
\vspace{-0.2cm}
\caption{An overview of our method. In GCM, $G_{x}$ uses $x$ from domain $X$ for training. In GAM, $G_{y}$ uses $y$ from domain $Y$ for training. Compared to $G_{x}$, $G_{y}$ has an extra encoder ($E_{r}$) to extract a feature $r$~(marked with a red box) as input. We use $G_{x}$ to get the corrected result $\tilde y^{x}$ which is used for training $G_{y}$~(Synthesis-As-Training Method). For PAM, $G_{pre}$ is trained using the left~(or right) eye $y^{l}_{l}$ and the flipping result $F(y^{r}_{l})$ of the right~(or left) eye $y^{r}_{l}$ to learn angle-invariant content features $c$~(marked with a blue box). The encoder of $G_{pre}$ is used in dual in-painting models to extract the feature $r$ as an extra input. We then compute pixel-wise differences with the $L1$ loss to optimize $G$. Note that the adversarial loss and the reconstruction loss are also used, but they are not shown in this figure.}
\label{fig:model}
\end{figure*}

\section{Related Work}
{\bfseries Generative Adversarial Networks:}
Generative Adversarial Networks~\cite{goodfellow2014generative} are powerful generative models which learn a  distribution that mimics a given target distribution. They have been applied to many fields, such as low-level image processing tasks~(image in-painting~\cite{pathak2016context,IizukaSIGGRAPH2017}, image super-resolution~\cite{Ledig_2017_CVPR,DRIT,wang2018esrgan}), high-level semantic and style transfer~(image translation~\cite{isola2017image,tang2019local,Zhu_2017_ICCV,liu2017face,tang2019multichannel,park2020contrastive,mallya2020world}, person image synthesis~\cite{tang2020xinggan,tang2019cycle}, image manipulation~\cite{park2020swapping}).

{\bfseries Image Inpainting:}
Image inpainting, an important task in computer vision and graphics, aims at
filling the missing pixels of an image with plausibly synthetic content. Recently, CNN-based and GAN-based methods have shown promising performance on image inpainting~\cite{pathakCVPR16context,iizuka2017globally,liu2019coherent,zhang2018semantic}. Thus, some works try to apply an inpainting model for facial attribute manipulation, such as hair, mouth and eye~\cite{Jo_2019_ICCV,dolhansky2018eye,olszewski2020intuitive}. Similar to these methods, our approach also follows the deep inpainting methods, but it does not require the data to be labeled or other additional information, such as a semantic segmentation mask, a sketch and even a reference image.


{\bfseries Gaze Correction:}
Previous work for gaze correction can be divided into three classes: 1) hardware-driven,
2) rendering and synthesis, 3) learning-based.

The hardware support is indispensable in early research. Kollarits~\etal~\cite{kollarits199634} tried to make use of half-silvered mirrors to allow
the camera to be placed on the optical path of the display. Yang~\etal~\cite{yang2004eye} aimed to address the eye contact problem with a novel
view synthesis, and they use a pair of calibrated stereo cameras and a face model to track the head pose in 3D. Generally speaking, these hardware-based methods are expensive.

Some works render the eye region based on a 3D fitting model, which replaces the original eye with
 new synthetic eyeballs. Banf~\etal~\cite{banf2009example} uses an example-based approach for deforming the eyelids and slide the iris across
the model surface with texture-coordinate interpolation. To fix the limitation caused by the use of a mesh, where the face and eyes are joined, GazeDirector~\cite{wood2018gazedirector}
regards the face and eyeballs as separate parts, synthesizing more high-quality images, especially for large redirection angles. These methods  can be applied to eye correction, but they struggle  when rendering eyes realistically is a challenge. Additionally, modeling methods have strong assumptions that usually do not hold in reality.

The core idea for most of the learning-based methods is to use a large paired training dataset to train the model~\cite{kononenko2015learning,kononenko2017photorealistic,yu2019improving,park2019few}. Some methods~\cite{kononenko2015learning,kononenko2017photorealistic} learn to generate the flow field which is used to relocate the eye pixels in the original image. Ganin~\etal~\cite{ganin2016deepwarp}
proposes to use a convolutional network to learn the flow field  warping the input image for redirecting the gaze to the desired angle. However, the model fails to generate photo-realistic and natural shape results of gaze redirection, as it uses only pixel-wise differences between the input and the ground truth as the loss. To solve this problem, He \etal~\cite{he2019photo} propose to use adversarial learning jointly with a cycle-consistent loss, which can improve the visual quality and redirection precision. However, these methods can hardly generate plausible results in the wild, coping with large variations in head pose, gaze angle and illumination. In contrast, we propose to utilize an in-painting model to correct the gaze angle which can generate  high-quality gaze correction and gaze animation in the wild.

\section{Method}
The overview of our method is shown in Fig.~\ref{fig:model} and it consists of three modules: the Gaze correction Module~(GCM), the Gaze Animation Module~(GAM) and the Pretrained Autoencoder Module~(PAM). Before introducing the details, we first clarify the notation.

$\bullet$ $z \in Z$ indicates an image instance represented in $Z = R^{m \times n \times c}$, where $m,n,c$ are the height, the width and the number of channels.

$\bullet$ $Z$ is divided into two domains: $X$  containing image instances with gaze staring at the camera and $Y$ containing image instance with gaze staring somewhere else.

$\bullet$ $M \in R^{m \times n \times c}$ denotes a binary mask of the eye region and $M^{'}$ defines the operation of extracting a sub-image~(eye region) for a rectangular region.

$\bullet$ $P_{x}, P_{y}$ denote distributions of data from the domains $X$ and $Y$, respectively.  $P_{m}$ denotes the distribution of data $M(z)$ with the eye regions removed. $M(x)$ and $M(y)$ have the same distribution, as the images $x \in X$ and $y \in Y$ have the only discrepancy in the eye region. Thus, $M(x) \thicksim P_{m}, M(y) \thicksim P_{m}$.

$\bullet$ $r \in R^{2}, c \in R^{256}$ denote the angle and the content features~(angle-invariant) where both are extracted by different encoders.

$\bullet$ $F$ denotes the image horizontal flipping operation (Mirroring).

\subsection{GCM with Generator $G_{x}$}

As shown in Fig.~\ref{fig:model}, $G_{x}$ is trained on data $x$ from domain $X$ and it aims to complete the masked image $x_{m} = M(x)$ by synthesizing the eye region.
In theory, $G_{x}$ can learn the mapping from $P_{m}$ to $P_{x}$, thus, $\tilde x = G_{x}(M(x))$ and $\tilde y^{x} = G_{x}(M(y))$. Then, $\tilde x \thicksim P_{x}$, $\tilde y^{x} \thicksim P_{x}$, as $M(x)$ and $M(y)$ have the same distribution $P_{m}$. This is the theoretical basis of our correction module. Specifically, we can use $G_{x}$ to correct the gaze to stare at the camera for data $y$ from domain $Y$. It can be represented as:
\begin{equation}
\begin{aligned}
& c_{x} = E_{c}(M^{'}(x)),  c_{y} = E_{c}(M^{'}(y)) \\
& \tilde x = G_{x}(M(x), c_{x}),  \tilde y^{x} = G_{x}(M(y), c_{y}),
\end{aligned}
\end{equation}
where $c_{x}$ and $c_{y}$ are the content features encoded from $M^{'}(x)$ and $M^{'}(y)$ by $E_{c}$. $E_{c}$ is the encoder of $G_{pre}$ which will be introduced in Section~\ref{PAM}.

We use a pixel-wise  loss ($L1$) for training GCM, shown in Fig.~\ref{fig:model}. It is defined as:
\begin{equation} \label{recon_loss_x}
\begin{aligned}
\ell^{x}_{recon} = \mathbb{E}_{x \thicksim P_{x}}[\Vert x - \tilde x \Vert_{1}].
\end{aligned}
\end{equation}

\begin{figure*}[ht]
\begin{center}
\includegraphics[width=0.9\linewidth]{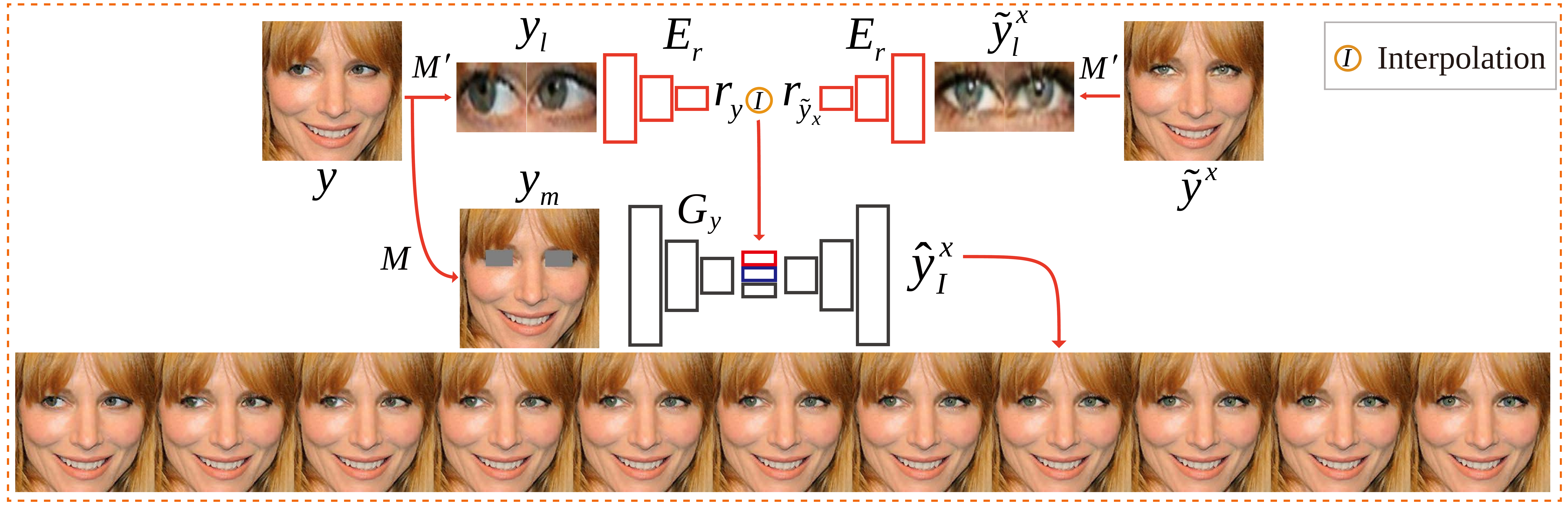}
\end{center}
\vspace{-0.5cm}
\caption{An overview of the architectures of our networks for gaze animation. $r_{y}$ and $r_{\tilde y^{x}}$ are the angle representation of the input image $y$ and the corrected result $\tilde y^{x}$, respectively. The bottom figure shows the interpolation process between $r_{y}$ and $r_{\tilde y_{x}}$, where the new angle representation is fed to $G_{y}$ to obtain the final output $\hat y^{x}_{I}$.}
\label{fig:model2}
\end{figure*}

\subsection{GAM with Generator $G_{y}$}

\begin{figure}[t]
\begin{center}
\includegraphics[width=0.9\linewidth]{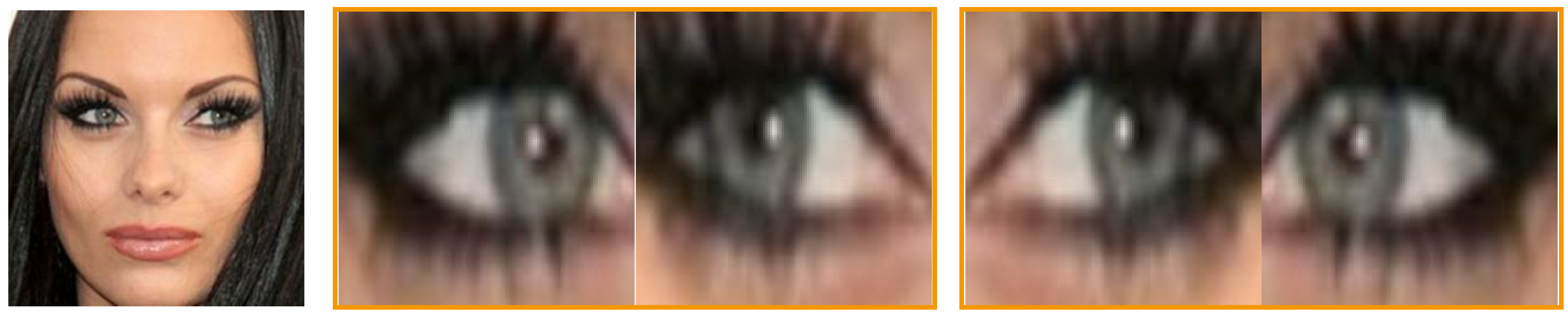}
\end{center}
\caption{Training samples in PAM. The first column is the input image $y$. The other columns show (from left to right): the left eye ($y^{l}_{l}$), the flipping result of right eye ($F(y^{r}_{l})$), the right eye ($y^{r}_{l}$), and the flipping result of left eye ($F(y^{l}_{l})$).}
\label{fig:model3}
\end{figure}

In addition to correct the gaze to stare at the camera, it is more valuable and efficient to redirect the gaze into any direction for gaze animation. Thus, we propose a novel Synthesis-As-Training Method, in which we use the synthetically corrected data as training data of the other generator $G_{y}$ to learn gaze animation.

In detail, this module can be divided into two stages. In the first stage we train the inpainting model $G_{y}$ to fill-in the masked image $y_{m} = M(y)$ and produce $\tilde y$. Different from $G_{x}$, $G_{y}$ encodes the eye region into the latent code $r \in R^{2}$ by exploiting $E_{r}$, where $r$ is an extra input of the decoder in $G_{y}$.
\begin{equation}
\begin{aligned}
r_{y} = E_{r}(M^{'}(y)), c_{y} = E_{c}(M^{'}(y)) \\
\tilde y = G_{y \thicksim P_{y}}(M(y), r_{y}, c_{y}).
\end{aligned}
\end{equation}

The reconstruction loss for PAM is defined as:
\begin{equation} \label{recon_loss_y}
\begin{aligned}
\ell^{y}_{recon} = \mathbb{E}_{y \thicksim P_{y}}[\Vert y - \tilde y \Vert_{1}].
\end{aligned}
\end{equation}

In the second stage, we use $G_{x}$ to correct the gaze of $y$ and produce the synthetic sample $\tilde y^{x}$ with the corrected gaze, shown in the right side of Fig.~\ref{fig:model}. Then, $\tilde y^{x}$ is used for training $G_{y}$, just like $y$ does. With the paired samples $(y, \tilde y^{x})$, which have the same incomplete region $M(y)$ but different eye region~(\ie, $y_{l}$, $\tilde y^{x}_{l}$), we train GAM and we ensure that the encoded feature $r$ has a highly correlation with the gaze angle.
\begin{equation}
\begin{aligned}
r_{\tilde y^{x}} = E_{r}(M^{'}(\tilde y^{x})), c_{\tilde y^{x}} = E_{c}(M^{'}(\tilde y^{x})) \\
\hat y = G_{y}(M(\tilde y^{x}), r_{\tilde y^{x}}, c_{\tilde y^{x}}).
\end{aligned}
\end{equation}

The corresponding loss is defined as:
\begin{equation} \label{recon_loss_y_x}
\begin{aligned}
\ell^{\tilde y^{x}}_{recon} = \mathbb{E}_{y \thicksim P_{y}}[\Vert \tilde y^{x} - \hat y \Vert_{1}].
\end{aligned}
\end{equation}

We exploit $\ell^{y}_{recon}$ + $\ell^{\tilde y^{x}}_{recon}$ as objective functions for training $E_{r}$ and $G_{y}$. And we interpolate this feature $r$ between $E_{r}(y_{l})$ and $E_{r}(\tilde y^{x}_{l})$ to produce the gaze animation which will be introduced later.

\subsection{PAM with Self-Supervised Mirror Learning} \label{PAM}

This inpainting process has difficulties in preserving the consistency of the identity information~(\eg, iris color, eye shape). Consequently, we propose to pretrain $G_{pre}$ to learn the content features ($c$). Then,  $c$ is used as the guidance for $G_{x}$ and $G_{y}$ to preserve the identity information of the inpainted results~($c$ denoted with a blue box in Fig.~\ref{fig:model}).


Although our training dataset is collected from the Internet, the majority of the images it is composed of, are close to the frontal image~(first column in Fig.~\ref{fig:model3}). As shown in columns 2-5 of Fig.~\ref{fig:model3}, we have paired images: left eye $y^{l}_{l}$ and the mirror $F(y^{r}_{l})$ of the right eye $y^{r}_{l}$, where both have different angles, but they have a very similar eye shape, iris color, etc., likewise for $y^{r}_{l}$. Based on this observation, as shown in the top-right of Fig.~\ref{fig:model}, we use two paired samples to pretrain $G_{pre}$ using the following objective functions:

\begin{equation} \label{recon_model}
\begin{aligned}
\ell_{pre} &=& \mathbb{E}_{y \thicksim P_{y}}[ \Vert y^{l}_{l} - G_{pre}(y^{l}_{l}) \Vert_{1}
+ \Vert y^{l}_{l} - G_{pre}(F(y^{r}_{l}) \Vert_{1}
\nonumber \\
&+& \Vert y^{r}_{l} - G_{pre}(y^{r}_{l}) \Vert_{1}
+ \Vert y^{r}_{l} - G_{pre}(F(y^{l}_{l})) \Vert_{1}
].
\end{aligned}
\end{equation}

After training, we found that the bottleneck feature $c$ of $G_{pre}$ is quasi angle-invariant, containing only the content information~(iris color, eye shape). Thus, we use the encoder network $E_{c}$ of $G_{pre}$ as extra input of $G_{x}$ and $G_{y}$ to extract content features. With these features as guidance, the inpainted results are more consistent with the input in identity information.

\subsection{Global and Local Discriminators for Adversarial Learning}
Since the $L_1$ loss tends to produce blurry results, we use two discriminators $D_{x}$ and $D_{y}$ adversarially trained.  $D_{x}$, $D_{y}$ do not share the weights of all the layers. Moreover, inspired by~\cite{iizuka2017globally}, we use a global discriminator that takes the entire face as input and a local discriminator which takes the local eye region as input. The global part is used to make coherent the entire image as a whole, while the local part is used to make the local region more realistic and  sharper. We concatenate the final fully-connected feature map of both parts into one output which is used as the input of a sigmoid function to predict the probability of the image being real.
The objective function for $D_{x}$ and $G_{x}$ is defined as:
\begin{eqnarray}
\mathop{min}\limits_{G_{x}}\mathop{max}\limits_{D_{x}}\ell^{x}_{adv} &=& \mathbb{\mathbb{E}}_{x \thicksim P_{x}}[logD_{x}(x, M^{'}(x))]
\nonumber \\ &+& \mathbb{\mathbb{E}}_{x \thicksim P_{x}}[log(1-D_{x}(\tilde x, M^{'}(\tilde x)))] \nonumber \\ &+& \mathbb{\mathbb{E}}_{y \thicksim P_{y}}[log(1 - D_{x}(\hat y, M^{'}(\hat y)))].
\label{eq_gan_loss_x}
\end{eqnarray}

The objective function for $D_{y}$ and $G_{y}$ is defined as:
\begin{eqnarray}
\mathop{min}\limits_{G_{y}}\mathop{max}\limits_{D_{y}}\ell^{y}_{adv} &=& \mathbb{\mathbb{E}}_{y \thicksim P_{y}}[logD_{y}(y, M^{'}(y))] \nonumber \\ &+& \mathbb{\mathbb{E}}_{y \thicksim P_{y}}[log(1-D_{y}(\tilde y, M^{'}(\tilde y)))].
\label{eq_gan_loss_y}
\end{eqnarray}

\subsection{Overall Loss}

Similar to~\cite{huang2018multimodal}, we use a reconstruction loss for the content features in the latent space to further preserve the identity information between the input image and the corrected output. This  loss ($L_{fp}$) is defined as:
\begin{equation}
\begin{aligned}
\ell_{fp} = \mathbb{E}_{y \thicksim P_{y}}[\Vert r_{y} - E_{r}(M^{'}(\tilde y)) \Vert_{1} + \Vert r_{\tilde y^{x}} - E_{r}(M^{'}(\hat y^{x})) \Vert_{1}].
\end{aligned}
\end{equation}

We use $- \ell^{x}_{adv}$ to train $D_{x}$ and $- \ell^{y}_{adv}$ to train $D_{y}$.
Concerning $G_{x}$, the overall loss is defined as:
\begin{eqnarray}
\ell^{x}_{all} = \ell^{x}_{adv} + \lambda_{1} \ell^{x}_{recon}.
\end{eqnarray}

For $G_{y}$ and $E_{r}$, the overall loss is defined as:
\begin{eqnarray}
\ell^{y}_{all} = \ell^{y}_{adv} + \lambda_{2} \ell^{x}_{adv} + \lambda_{3} \ell^{y}_{recon} + \lambda_{4} \ell^{\tilde y^{x}}_{recon} + \lambda_{5} \ell_{fp},
\end{eqnarray}
where $\lambda_{1}, \lambda_{2}, \lambda_{3}, \lambda_{4}$ and $\lambda_{5}$ are hyper-parameters controlling the contributions of each loss term.

\subsection{Inference for Gaze correction and Animation}

We obtain the correction result $\tilde y^{x}$ for a given sample $y$ using $G_{x}$. As shown in   Fig.~\ref{fig:model2} (top), we modify the encoded representation $r$ by interpolating between $r_{y}$ and $r_{\tilde y_{x}}$, which are encoded from the eye region $y_{l}$ and $\tilde y^{x}_{l}$, respectively. The new representation can be fed to $G_{y}$ to obtain animation results $\hat y^{x}_{I}$ with different angles. As shown in the bottom of Fig.~\ref{fig:model2}, we obtain plausible and smooth gaze animations in the wild.

\section{Experiments}

\begin{figure*}[t]
\begin{center}
\includegraphics[width=1.0\linewidth]{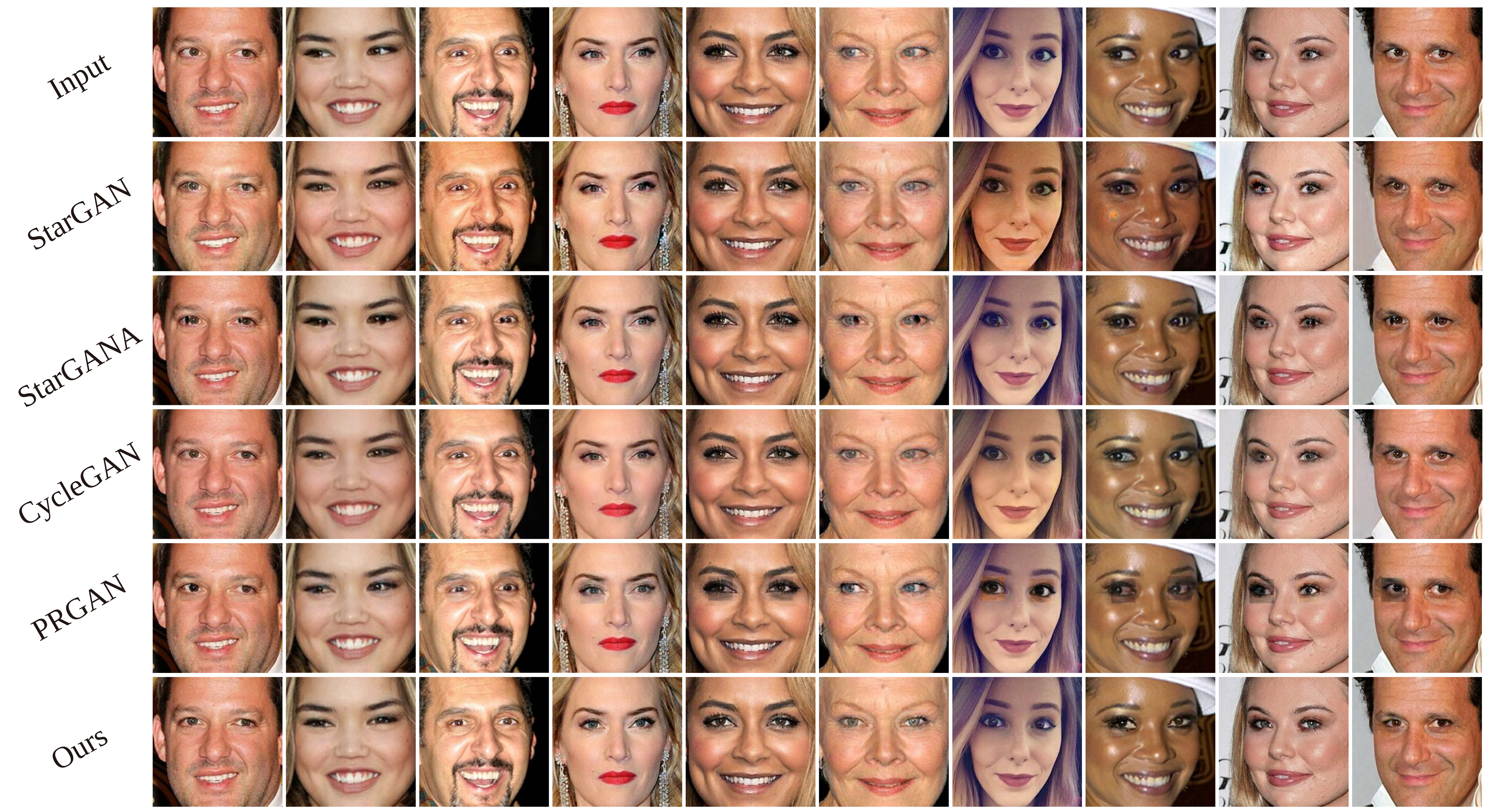}
\end{center}
\caption{Qualitative comparison for the gaze correction task. The first row shows the input images and the following rows show the gaze correction results of StarGAN~\cite{Choi_2018_CVPR}, StarGANA, CycleGAN~\cite{Zhu_2017_ICCV}, PRGAN~\cite{he2019photo} and GazeGAN.}
\label{fig:exp1}
\end{figure*}

In this section, we first introduce the details of our dataset, the network training and the baseline models.
Then, we compare the proposed method with the state-of-the-art methods on gaze correction in the wild using both qualitative and quantitative evaluations.
Next, we demonstrate the effectiveness of the proposed method on gaze animation with various outputs by interpolating and extrapolating in the latent space.
Finally, we present an ablation study to validate the effect of each component of our model, \ie, Synthesis-As-Training Method,
Pretrained Autoencoder with Self-Supervised Mirror Learning, and Latent Reconstruction Loss. For brevity, we refer to our full method as~\textbf{GazeGAN}. {\em Note that we do not use any post-processing algorithm for GazeGAN.}

\subsection{Dataset}
{\bfseries CelebAGaze Dataset:}
Most of the existing benchmark datasets~\cite{funes2014eyediap,smith2013gaze,zhang2017mpiigaze, zhang2020eth} are not suitable for our gaze correction task in the wild, which asks for a wider gaze range, various head poses and different illumination conditions. Recently,~\cite{kellnhofer2019gaze360} presented a large scale gaze tracking dataset, called Gaze360, for robust 3D gaze estimation in unconstrained images. Although this dataset has been labeled with a 3D gaze with a wide range of eye angles and head poses, it still lacks high-resolution images for face and eye regions. Additionally, this dataset does not provide the eye data with staring at the camera, which is required by our proposed unsupervised method.

To remedy this problem, we collected a new dataset, called CelebAGaze. In detail, CelebAGaze consists of 25283 high-resolution celebrity images that are collected from CelebA~\cite{liu2015deep} and the Internet.
It consists of 21832 face images with eyes staring at the camera and 3451 face images with eyes staring  somewhere else. We cropped all images~($256 \times 256$) and compute the eye mask region by dlib~\cite{king2009dlib}. Specifically, we use dlib to extract 68 facial landmarks and calculate the mean of 6 points near the eye region, which will be the center point of the mask. The size of the mask is fixed to $30\times 50$. As described above, we randomly select 300 samples from domain $Y$, 100 samples from domain $X$ as the test set, the remaining as the training set. Note that this dataset is unpaired and it is not labeled with the specific eye angle or the head pose information.
We show some samples of the CelebAGaze dataset in Fig.~\ref{fig:newgaze}. 

\subsection{Training Details}
We first train the PAM module. Then, the discriminators  $D_{x}$ and $D_{y}$ and the generators  $G_{x}$ and $G_{y}$ are jointly optimized. We use the Adam optimizer with $\beta_{1}=0.5$ and $\beta_{2}=0.999$. The batch size is 8. The initial learning rate for PAM is 0.0005 and 0.0001 for the discriminators   and the generators in the first 20000 iterations, and linearly decayed to 0 over the remaining iterations. All the weight coefficients $\lambda_{1}, \lambda_{2}, \lambda_{3}, \lambda_{4}$, $\lambda_{5}$ are set to 1. To stabilize the network training in the adversarial learning, we use  spectral normalization~\cite{miyato2018spectral} for all the convolutional layers of the discriminators $D_{x}$ and $D_{y}$, but not for the generator $G_{x}$. Our model is implemented in Tensorflow and takes ten hours to be trained with a single NVIDIA Titan X GPU.

\subsection{Baseline Models}
{\bfseries Gaze Correction:}
PRGAN~\cite{he2019photo} achieved state-of-the-art gaze redirection results on the Columbia dataset based on a single encoder-decoder network with  adversarial learning, similarly to StarGAN~\cite{Choi_2018_CVPR}. The original PRGAN is trained on paired samples with labeled  angles~(Columbia Gaze~\cite{smith2013gaze}). To train PRGAN on the proposed CelebAGaze dataset, we remove the VGG perceptual loss of PRGAN and learn the domain translation between $X$ and $Y$. Note that we train PRGAN only with the local eye region, the same way as in the original paper.

\begin{table}
\vspace{-0.1cm}
\caption{Columns 1-2: the MSSSIM and LPIPS scores computed on the background regions of the gaze correction results using different models. Column 3: the user study results. Higher is better for MSSSIM and the user study; lower is better for LPIPS.}
\begin{center}
\small
\begin{tabular}{ccccc}
\hline
Metrics & MSSSIM $\uparrow$ & LPIPS $\downarrow$  & User Studies $\uparrow$  \\
\hline
Other   & - & -  & 24.20\% \\
StarGAN~\cite{Choi_2018_CVPR} & 0.969 & 0.0732 & 3.40 \% \\
StarGANA & 0.998 & 0.0022 & 6.67 \% \\
CycleGAN~\cite{Zhu_2017_ICCV} & 0.991 & 0.0267 & 15.00 \% \\
PRGAN~\cite{he2019photo} & {\bfseries 1.0}  & {\bfseries 0.0} & 8.33 \% \\
GazeGAN & {\bfseries 1.0}  & {\bfseries 0.0} & {\bfseries 42.40 \%} \\
\hline
GT  & 1.0 & 0.0 & 100\% \\
\end{tabular}
\label{tab:evaluate1}
\end{center}
\end{table}

{\bfseries Facial Attribute Manipulation:}
Gaze correction and animation can be regarded as a sub-task of facial attribute manipulation. Recently, StarGAN~\cite{Choi_2018_CVPR}  achieved very high-quality results in facial attribute manipulation. We train StarGAN on the CelebAGaze dataset to learn the translation mapping between domain $X$ and domain $Y$. To further improve the ability of StarGAN on this task, we employ a variant of StarGAN as an additional baseline: StarGAN with spatial attention learning~\cite{zhang2018generative} referred to as StarGANA. Moreover, gaze correction can be considered as an image translation task. Thus, we adopt CycleGAN as another baseline for our experiments. Note that we do not compare GazeGAN with SGGAN~\cite{Zhang:2018:SGM:3240508.3240594}, AttGAN~\cite{he2019attgan}, STGAN~\cite{liu2019stgan}, RelGAN~\cite{wu2019relgan} and PAGAN~\cite{he2020pa}, as they have a performance  very close  to StarGAN in facial attribute manipulation. We use the public code of StarGAN~\footnote{\url{https://github.com/yunjey/StarGAN}}, CycleGAN~\footnote{\url{https://github.com/junyanz/pytorch-CycleGAN-and-pix2pix}} and PRGAN~\footnote{\url{https://github.com/HzDmS/gaze\_redirection}}.

\subsection{Gaze correction}

In this section, we first qualitatively compare the proposed method with state-of-the-art methods on the CelebAGaze dataset for the task of gaze correction. Then, we choose some metrics to quantitatively evaluate the gaze correction results.

\begin{figure*}[t]
\begin{center}
\includegraphics[width=1.0\linewidth]{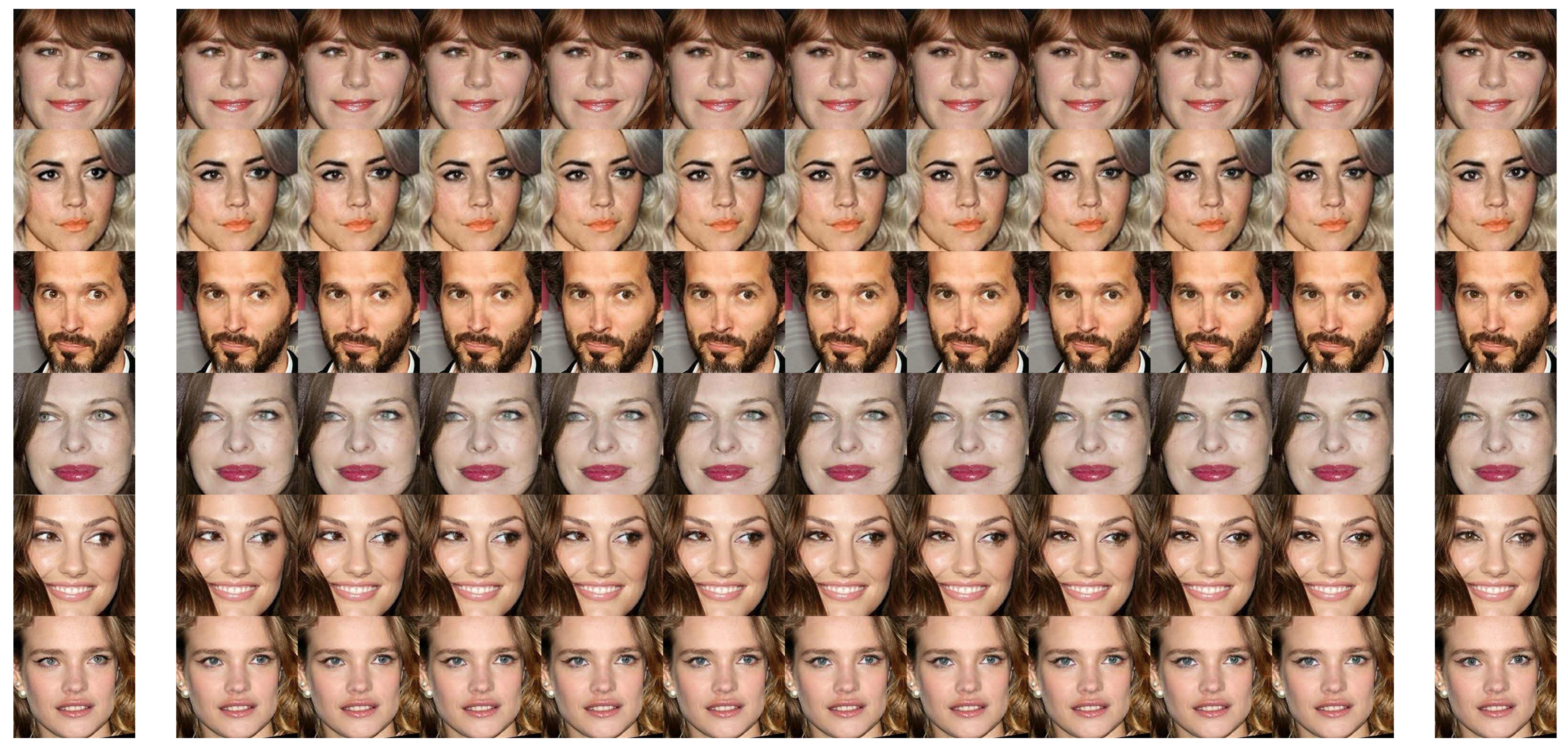}
\end{center}
\caption{Gaze animation results using the interpolation of the latent features $r$. The first and the final column show the input images and corrected results, respectively. The middle columns show the interpolated images.}
\label{fig:exp2}
\vspace{-0.3cm}
\end{figure*}

\begin{figure}[t]
\begin{center}
\includegraphics[width=1.0\linewidth]{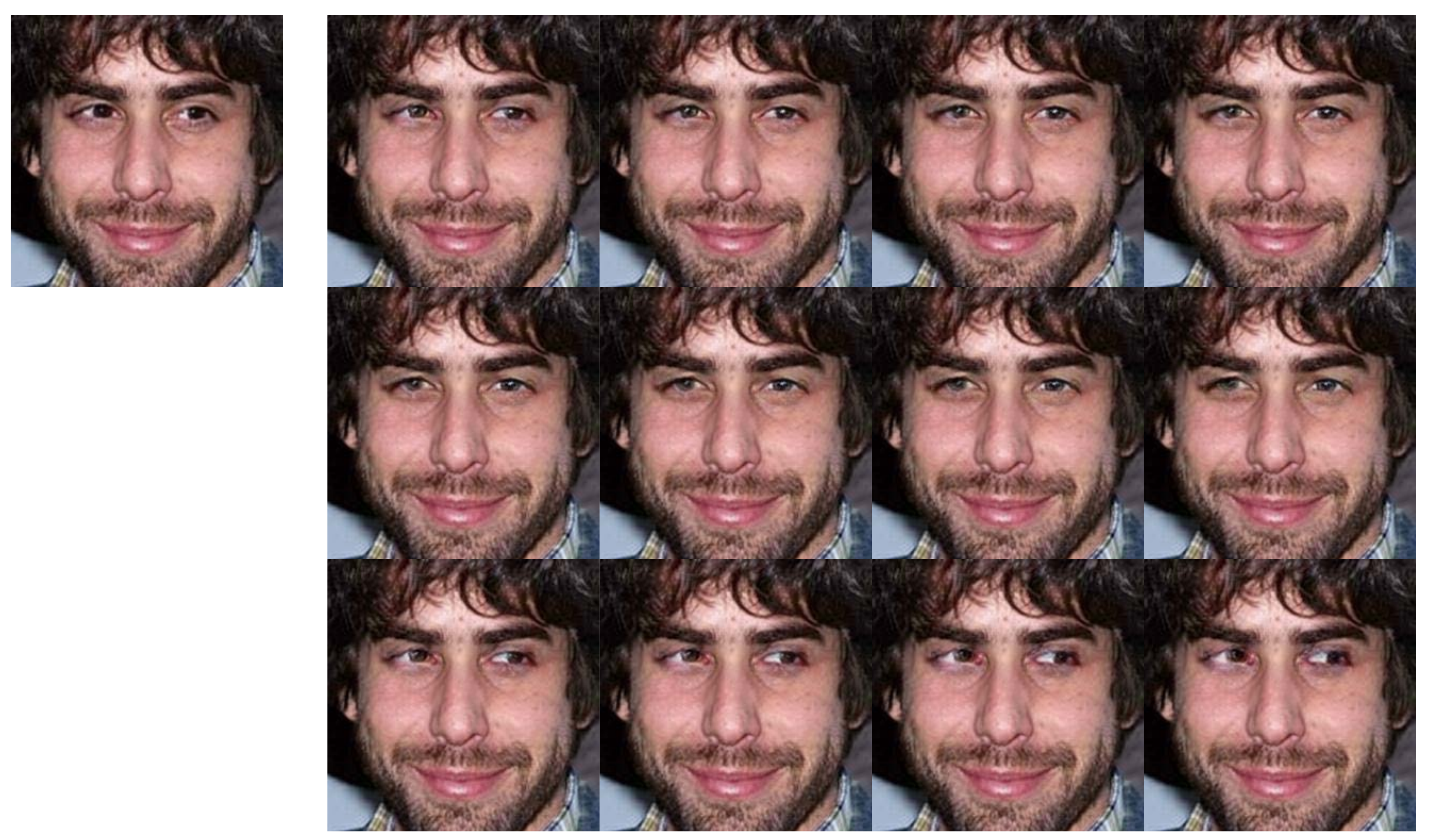}
\end{center}
\caption{Gaze animation examples obtained by both  interpolation and extrapolation of the latent features $r$. The $1st$ row shows the interpolation results and the next two rows show the extrapolation results.}
\label{fig:exp3}
\end{figure}

{\bfseries Qualitative results:}
As shown in the last row of Fig.~\ref{fig:exp1},  GazeGAN can correct the eyes to look at the camera while preserving the identity information such as the eye shape and the iris color, validating the effectiveness of the proposed method. The $2nd$ row shows the results of StarGAN~\cite{Choi_2018_CVPR}. We note that StarGAN could not produce precise gazes  staring at the camera and it suffers for the low-quality generation with lots of artifacts. The results of CycleGAN are shown in $4th$ row. Although the results of CycleGAN are very realistic and with low artifacts in the eye region, this method does not produce a precise correction for the gaze. We believe that this is because both StarGAN and CycleGAN are based on the cycle-consistency loss, which requires that the mapping between $X$ and $Y$ should be continuous and invertible. According to the invariance of the Domain Theorem~\footnote{\url{https://en.wikipedia.org/wiki/Invariance\_of\_domain}}, the intrinsic dimensions of the two domains should be the same. However, the intrinsic dimension of $Y$ is much larger than $X$, as $Y$ has more variations for the gaze angle than $X$. As shown in the $7th$ columns of Fig.~\ref{fig:exp1}, both StarGAN and CycleGAN cannot preserve the irrelevant regions of the face.
Another stronger baseline is StarGANA, which combines  spatial attention~\cite{zhang2018generative} with StarGAN, and successfully preserves the irrelevant region~(shown in the $3rd$ row).
However, the results of StarGANA are not faithfully and realistic, with lots of deformation artifacts, compared with GazeGAN.

Moreover, we compare GazeGAN with PRGAN~\cite{he2019photo}, which is trained  using only local eye regions (same as in the original paper), which may be helpful to focus on the translation in the eye region. The results of PRGAN are shown in the $5th$ row of Fig.~\ref{fig:exp1}. Compared with GazeGAN, PRGAN does not produce precise and realistic correction results, similarly to StarGAN and CycleGAN. Additionally, PRGAN suffers from the boundary mismatch problem between the local eye region and the global face.


\begin{figure*}[t]
\begin{center}
\includegraphics[width=1.0\linewidth]{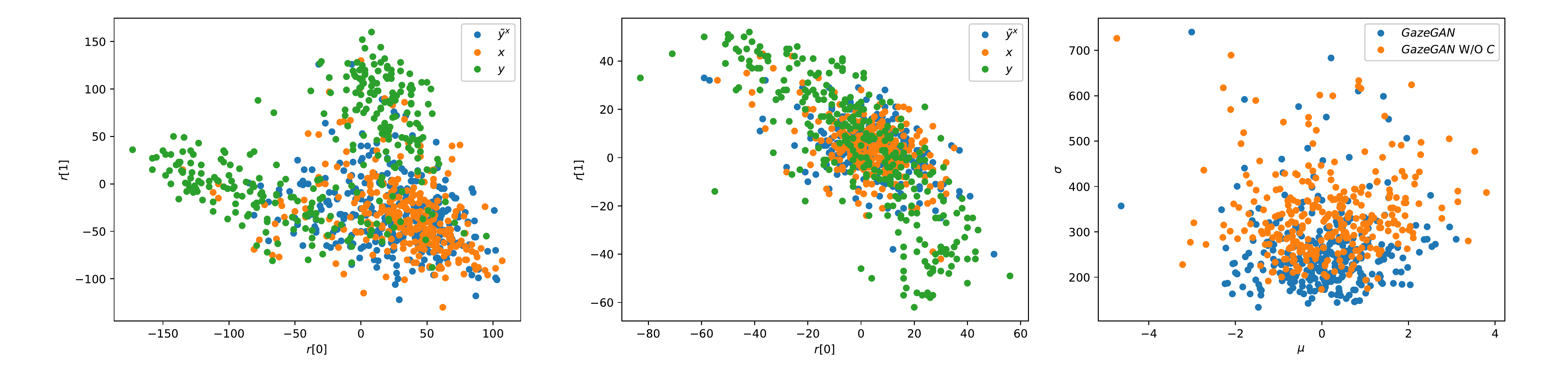}
\end{center}
\vspace{-0.5cm}
\caption{A visualization  of the angle~(columns 1-2) and the content~(column 3) representations. Columns 1-2 correspond to the results of GazeGAN W/O $B$ and GazeGAN, respectively. $r[0]$ and $r[1]$ mean the $1st$ and $2nd$ value of the $r$ feature vector, respectively. Column 3 shows the mean~($X$-coordinate) and the variance~($Y$-coordinate) of the content representation differences between GazeGAN W/O $C$ and GazeGAN.}
\label{fig:exp4}
\vspace{-0.3cm}
\end{figure*}

\begin{figure}[t]
\begin{center}
\includegraphics[width=1.0\linewidth]{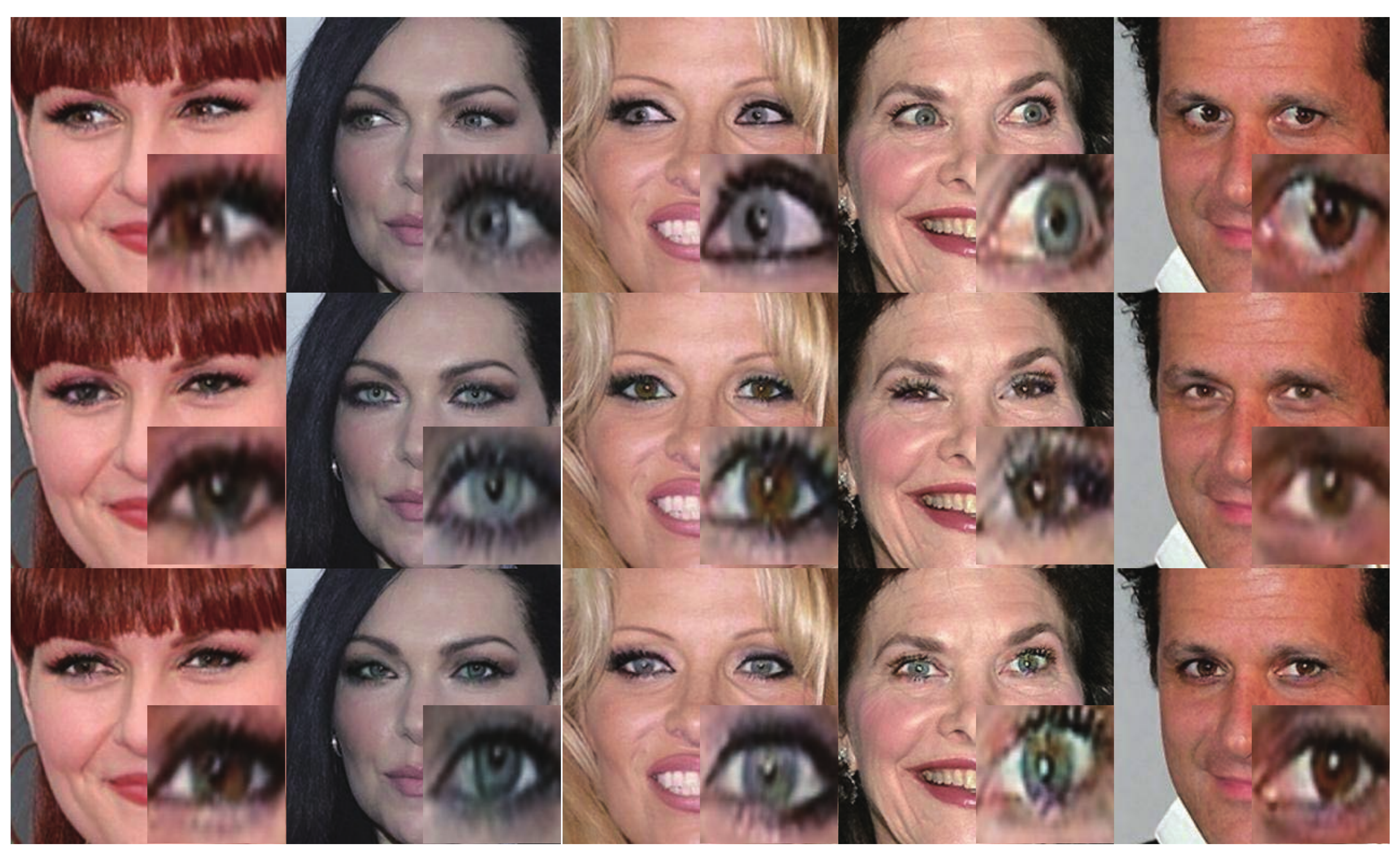}
\end{center}
\caption{A qualitative comparison between GazeGAN and GazeGAN W/O $A$. The first row shows the input images and next two rows are the corrected results of GazeGAN W/O $A$ and GazeGAN~(Zoomed left eyes).}
\label{fig:exp5}
\end{figure}

{\bfseries Quantitative Evaluation Protocol:}
The qualitative evaluation has validated the effectiveness and the superiority of our proposed GazeGAN in gaze correction. To further support the previous evaluation with a quantitative evaluation, we employ MSSSIM~\cite{wang2003multiscale} and LPIPS~\cite{zhang2018perceptual} to measure the {\em preservation} ability of the {\em irrelevant regions}, \ie, faces without the eye region ($M(y)$). Specifically, we compute the mean MSSSIM and LPIPS scores between $M(y)$ and $M(\tilde y)$ across all the test data from domain $Y$.

In addition to the metrics aforementioned, we conduct a user study to assess the
results of the gaze correction from different models. In detail, given an input face image in the CelebAGaze test dataset (extracted from $Y$), we show the corrected results from different models to 30 respondents who were
asked to select the best image based on perceptual realism and the precision of the gaze correction. They also can select ``Other'' which means that the results from all the models are not satisfactory. This study is based on 50 questions for all respondents.

\begin{table}
\vspace{-0.1cm}
\caption{Comparison between GazeGAN and GazeGAN W/O $A$, where the latter  denotes removing the content representation extracted from~$E_{c}$.
The scores are measured between the input image $x$ and inpainted result $\tilde x$ across all the test data from $X$.
}
\vspace{-0.2cm}
\begin{center}
\small
\begin{tabular}{ccc}
\hline
Metrics &  GazeGAN & GazeGAN W/O A \\
\hline
MSSSIM $\uparrow$  & {\bfseries 0.67} & 0.52 \\
LPIPS $\downarrow$  & {\bfseries 0.1680} & 0.2410 \\
\hline
\end{tabular}
\label{tab:exp2}
\vspace{-0.4cm}
\end{center}
\end{table}

\begin{table}
\vspace{-0.1cm}
\caption{
Comparison with GazeGAN W/O $C$, which denotes removing the latent reconstruction loss $\ell_{fp}$.
The scores are measured between the input image $y$ and the reconstruction result $\tilde y$ across all the test data from $Y$.
}
\vspace{-0.3cm}
\begin{center}
\small
\begin{tabular}{ccc}
\hline
Metrics &  GazeGAN & GazeGAN W/O $C$ \\
\hline
MSSSIM $\uparrow$  & {\bfseries 0.59} & 0.53 \\
LPIPS $\downarrow$  & {\bfseries 0.2506} & 0.2675 \\
\hline
\end{tabular}
\label{tab:exp3}
\vspace{-0.4cm}
\end{center}
\end{table}

{\bfseries Quantitative results:}
The first two columns of Table~\ref{tab:evaluate1} show the MSSSIM and LPIPS scores  evaluating the preservation ability of the corrected images across different models. GazeGAN and PRGAN obtain the best results with 1.0 for MSSSIM and 0.0 for LPIPS. In fact, the original irrelevant regions are integrated with the generated eye region in both models using binary masks. StarGAN, StarGANA and CycleGAN obtain the worse irrelevant region preservation  scores.
The last column of Table~\ref{tab:evaluate1} shows the evaluation results of the user study. The user average vote for GazeGAN is $42.40\%$, which is higher than all the other methods, \ie, $3.40\%$ higher than StarGAN, $6.67\%$ higher than StarGANA, $15.00\%$ higher than CycleGAN, $8.33\%$ higher than PRGAN. Overall, the qualitative and quantitative evaluations demonstrate the effectiveness and the superiority of the proposed approach.

\subsection{Gaze Animation}
In Fig.~\ref{fig:exp2} we show the  gaze animation results using input images with an arbitrary gaze which are corrected to stare at the camera. We  observe that the interpolation results are smooth and plausible in each row. Each column has a different gaze angle, but the identity information is overall preserved~(\eg, the eye shape, the iris color). Next, we show gaze animation results with more various directions by {\em extrapolating} the features $r$, in addition to interpolation methods. As shown in Fig.~\ref{fig:exp3}, our model not only achieves high-quality interpolation results but also achieves gaze animation with more directions where the angles are out of the range between the input and corrected output.

\subsection{Ablation Study}
In this section, we conduct extensive ablation studies to investigate the three key components of the proposed GazeGAN, \ie, the Pretrained Autoencoder for content extraction, the Synthesis-As-Training Method, and the Latent Reconstruction Loss $\ell_{fp}$. For brevity, we refer to these components as $A$, $B$ and $C$, respectively.

{\bfseries Pretrained Autoencoder:} We propose this pretrained model to extract the content features to guide the process of gaze correction while preserving the identity information.
We observe that GazeGAN has a stronger ability to preserve identity information with respect to GazeGAN W/O $A$ in Fig.~\ref{fig:exp5}.
To quantitatively evaluate this, we use $G_{x}$ to reconstruct the input image $x$ from the test dataset in the domain $X$ and we measure the differences between the input images and the correction results in the local eye regions by employing MSSSIM and LPIPS. Table~\ref{tab:exp2} shows that GazeGAN achieves better scores than GazeGAN W/O $A$, validating our design motivation.

{\bfseries Training-As-Synthesis Method:} The results in Fig.~\ref{fig:exp2} have shown the effectiveness of this method for gaze animation. Additionally, we illustrate three groups of features $r$ extracted from $E_{r}$ with real test samples $y$, real test samples $x$, the corrected results $\tilde y^{x}$ as input. In Fig.~\ref{fig:exp4}, the $1st$ and the $2nd$ column correspond to the results of GazeGAN W/O $B$, and  GazeGAN respectively.
We observe that the points corresponding to $\tilde y^{x}$ are closer to the points corresponding to  $x$ in the $2nd$ figure with respect to what happens in the $1st$ column, which indicates that features $r$ have a strong correlation with angle information.

{\bfseries Latent Reconstruction Loss $\ell_{fp}$:} We use $G_{y}$ to complete the input images $y$ from the test data of $Y$ and we measure the differences between the input images and the results using MSSSIM and LPIPS. In Fig.~\ref{tab:exp3}, GazeGAN obtains higher scores than GazeGAN W/O $C$, which shows that $\ell_{fp}$ further improves the ability to preserve identity information. Additionally, we visualize the mean $\mu$~($X$-coordinate) and the variance $\sigma$~($Y$-coordinate) of the arithmetical differences $c^{\tilde y} - c^{y}$ extracted from $G_{c}$ across all the test data from $Y$. As shown in the right of Fig.~\ref{fig:exp4}, we see that our full model GazeGAN has a lower variance than GazeGAN W/O $C$, as the loss encourages to reduce the differences of the content features.

\section{Conclusion}
In this paper we introduced a new gaze dataset in the wild, CelebAGaze, which is characterized by a large diversity in the head poses, gaze angles and illumination conditions. Moreover, we  presented a novel unsupervised model, GazeGAN, for gaze correction and animation, trained and tested on this dataset. GazeGAN generates more high-quality and precise gaze correction results than state-of-the-art methods. Furthermore, it can redirect the gaze into the desired direction, producing gaze animation by interpolation in the latent space. Future work includes extending the proposed model to gaze redirection tasks in videos.
\bibliographystyle{ACM-Reference-Format}
\bibliography{sample-base}

\clearpage

\appendix

\section{Introduction}

This supplementary document provides additional results supporting the claims of the main paper.

\begin{figure}[htp]
\section{Gaze Correction.}
\begin{center}
\includegraphics[width=1.0\linewidth]{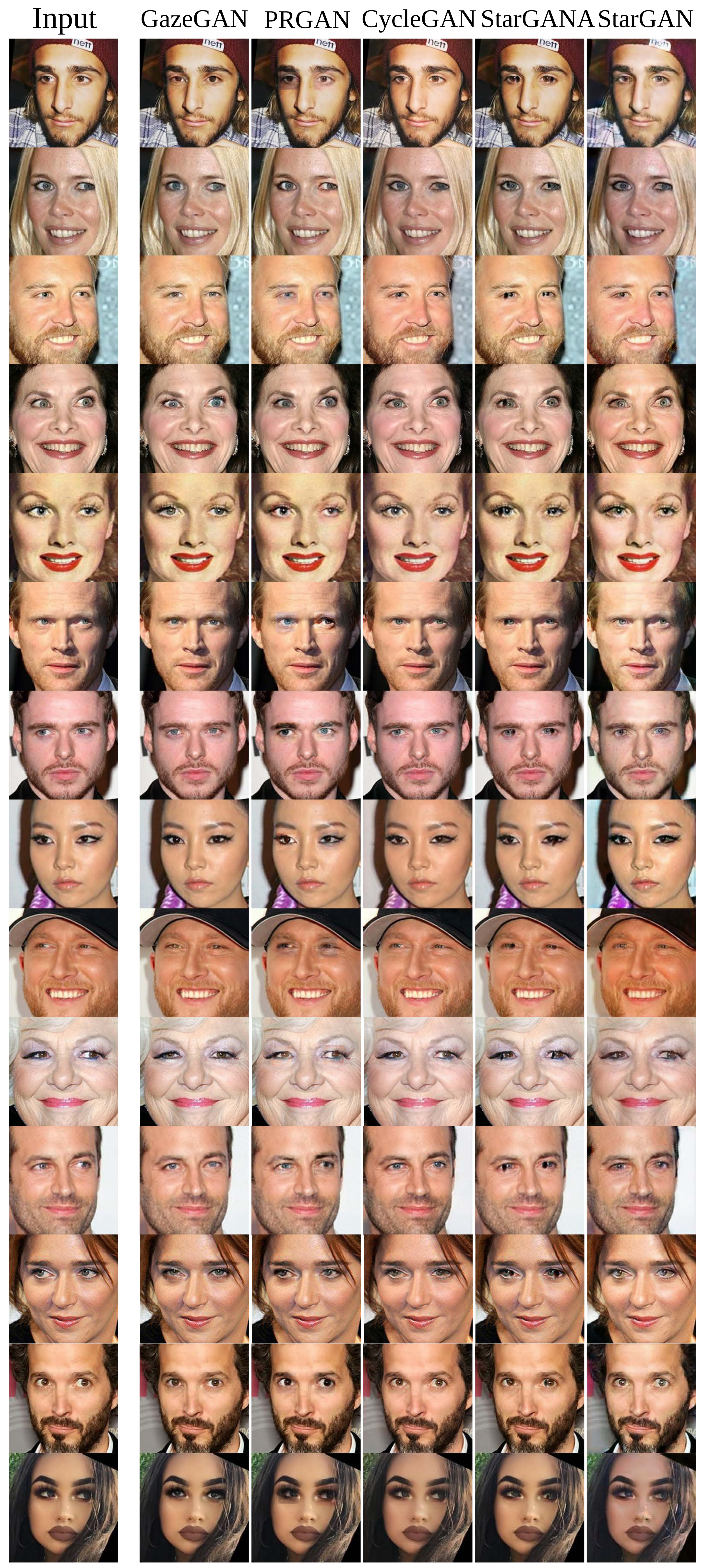}
\end{center}
\vspace{-0.3cm}
\caption{More comparison results for gaze correction in test data from domain $Y$. We can found GazeGAN achieves more realistic results with precise gaze angle than the baselines.}
\label{fig:correction}
\end{figure}

\begin{figure*}[hp]
\section{Gaze Animation.}
\begin{center}
\includegraphics[width=1.0\linewidth]{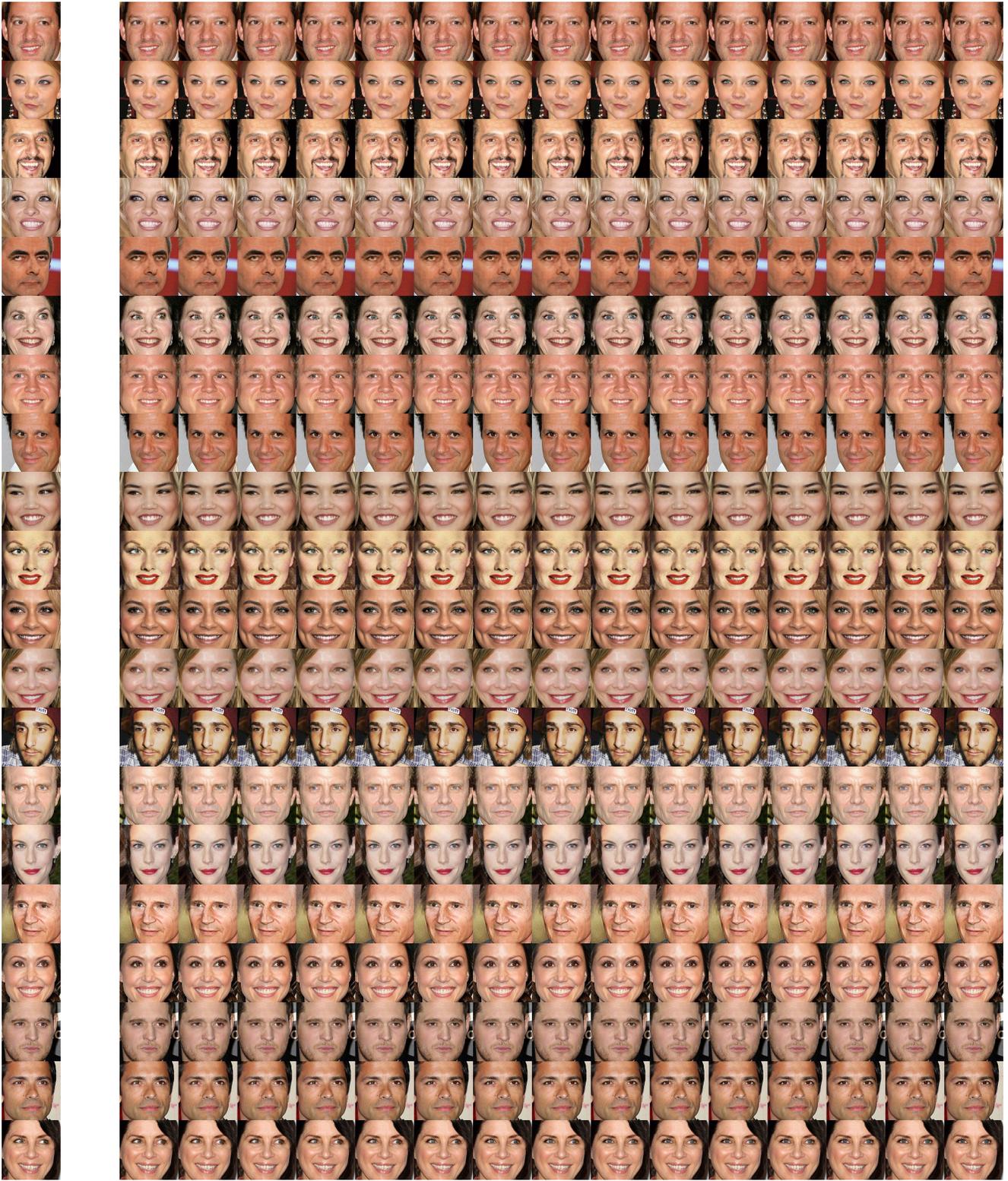}
\end{center}
\vspace{-0.3cm}
\caption{More gaze animation results in the wild.}
\label{fig:animation}
\end{figure*}

Firstly, we show the network architecture of $G_{x}$, $G_{y}$, $D$ and $G_{pre}$ in Table~\ref{tab:gx}, Table~\ref{tab:gy}, Table~\ref{tab:d} and Table~\ref{tab:gpre}. Note that $D_{x}$ and $D_{y}$ have the same architecture $D$. Secondly, we show more examples of CelebAGaze dataset in Fig.~\ref{fig:newgaze}. Finally, Fig.~\ref{fig:correction} and Fig.~\ref{fig:animation} show more gaze correction and animation results by interpolation in latent space to validate the effectiveness and superiority of GazeGAN.

\section{Network Architecture}

Here are some notations should be noted: CONV: convolutional layer; DECONV: deconvolutional layer. $h$: height of input images; $w$: width of the input images; C: number of output channels; K: size of kernels; S: strides of kernels; $IN$: instance normalization; LReLU: Leaky ReLU; In our experiments, $h=256, w=256$.

\begin{table}[htp]
\begin{center}
\begin{tabular}{|lll|}
\hline
Module & Input Shape  & Layer Information \\
\hline\hline
\multirow{6}*{Encoder} & $(h,w,3+1)$ & Input \\
& $(h,w,4)$ & CONV-(C16, K7$\times$7, S1$\times$1),$IN$,LReLU\\
& $(h,w,16)$ & CONV-(C32, K4$\times$4, S2$\times$2),$IN$,LReLU \\

& $(\frac{h}{2},\frac{w}{2},32)$ & CONV-(C64, K4$\times$4, S2$\times$2),$IN$,LReLU \\

& $(\frac{h}{4},\frac{w}{4},64)$ & CONV-(C128, K4$\times$4, S2$\times$2),$IN$,LReLU \\

& $(\frac{h}{8},\frac{w}{8},128)$ & CONV-(C256, K4$\times$4, S2$\times$2),$IN$,lRelu \\

& $(\frac{h}{16},\frac{w}{16},256)$ & CONV-(C256, K4$\times$4, S2$\times$2),$IN$,LReLU \\

& $(\frac{h}{32},\frac{w}{32},256)$ & FC-(C256)\\
\hline

\hline
\multirow{6}*{Encoder $E_{r}$} & $(\frac{h}{2},\frac{w}{2},3)$ &  Input \\
& $(\frac{h}{2},\frac{w}{2},3)$ & CONV-(C32, K7$\times$7, S1$\times$1),$IN$,LReLU \\

& $(\frac{h}{2},\frac{w}{2},32)$ & CONV-(C64, K4$\times$4, S2$\times$2),$IN$,LReLU \\

& $(\frac{h}{4},\frac{w}{4},64)$ & CONV-(C128, K4$\times$4, S2$\times$2),$IN$,LReLU \\

& $(\frac{h}{8},\frac{w}{8},128)$ & CONV-(C128, K4$\times$4, S2$\times$2),$IN$,LReLU \\

& $(\frac{h}{16},\frac{w}{16},128)$ & FC-(C2) \\
\hline

\hline
\multirow{7}*{Decoder} & (258) & FC-(C256 $\times \frac{h}{32} \times \frac{w}{32})$ \\

~ & $(\frac{h}{32},\frac{w}{32},256)$ &  DECONV-(C128,K4$\times$4,S2$\times$2),$IN$,LReLU \\

~&  $(\frac{h}{16},\frac{w}{16},128)$ &  DECONV-(C64,K4$\times$4,S2$\times$2 ),$IN$,LReLU \\

~&  $(\frac{h}{8},\frac{w}{8},64)$  &  DECONV-(C32,K4$\times$4,S2$\times$2),$IN$,LReLU \\

~&  $(\frac{h}{4},\frac{w}{4},32)$  &  DECONV-(C16,K4$\times$4,S2$\times$2),$IN$,LReLU \\

~&  $(\frac{h}{2},\frac{w}{2},16)$  &  DECONV-(C16,K4$\times$4,S2$\times$2),$IN$,LReLU  \\

~ & $(h,w,16)$ & CONV-(C3, k7$\times$7, S1$\times$1),Tanh \\

\hline
\end{tabular}
\end{center}
\vspace{0.3cm}
\caption{The architectures of Encoder and Decoder in generator $G_{y}$. The mask $M$ would be also as input of $G_{y}$.}
\label{tab:gy}
\end{table}

\renewcommand{\arraystretch}{1.2}

\begin{table}[htp]
\begin{center}
\begin{tabular}{|lll|}
\hline
Module & Input Shape  & Layer Information \\
\hline
\multirow{8}*{Global D} & ($h$,$w$,$3$) & Input \\
& ($h$,$w$,3) & CONV-(C32 K4$\times$4,S2$\times$2),LReLU \\

& ($\frac{h}{2}$,$\frac{w}{2}$,32) & CONV-(C64 K4$\times$4,S2$\times$2),LReLU \\

& ($\frac{h}{4}$,$\frac{w}{4}$,64) & CONV-(C128 K4$\times$4,S2$\times$2),LReLU  \\

& ($\frac{h}{8}$,$\frac{w}{8}$,128) & CONV-(C256 K4$\times$4,S2$\times$2),LReLU \\

& ($\frac{h}{16}$,$\frac{w}{16}$,256) & CONV-(C256 K4$\times$4,S2$\times$2),LReLU \\

& ($\frac{h}{32}$,$\frac{w}{32}$,256) & CONV-(C256 K4$\times$4,S2$\times$2),LReLU \\

~ & $(\frac{h}{64},\frac{w}{64},256)$ & FC-(C256) \\

\hline
\multirow{8}*{Local D} & ($\frac{h}{2}$,$\frac{w}{2}$,$3$) & Input \\
& ($\frac{h}{2}$,$\frac{w}{2}$,3) & CONV-(C32 K4$\times$4,S2$\times$2),LReLU \\

& ($\frac{h}{4}$,$\frac{w}{4}$,32) & CONV-(C64 K4$\times$4,S2$\times$2),LReLU \\

& ($\frac{h}{8}$,$\frac{w}{8}$,64) & CONV-(C128 K4$\times$4,S2$\times$2),LReLU  \\

& ($\frac{h}{16}$,$\frac{w}{16}$,128) & CONV-(C256 K4$\times$4,S2$\times$2),LReLU \\

& ($\frac{h}{32}$,$\frac{w}{32}$,256) & CONV-(C256 K4$\times$4,S2$\times$2),LReLU \\

& ($\frac{h}{64}$,$\frac{w}{64}$,256) & CONV-(C256 K4$\times$4,S2$\times$2),LReLU \\

~ & $(\frac{h}{128},\frac{w}{128},256)$ & FC-(C256) \\
\hline
\multirow{2}*{Concat} & (512) & FC-(C512),LReLU \\
& (512) & FC-(C1) \\
\hline
\end{tabular}
\end{center}
\caption{The architectures of global and local discriminator $D_{x}$ and $D_{y}$. Both have the same architecture.}
\label{tab:d}
\end{table}

\begin{table}[htp]
\begin{center}
\begin{tabular}{|lll|}
\hline
Module & Input Shape  & Layer Information \\
\hline\hline
\multirow{6}*{Encoder} & $(h,w,3+1)$ & Input \\
& $(h,w,4)$ & CONV-(C16, K7$\times$7, S1$\times$1),$IN$,LReLU\\
& $(h,w,16)$ & CONV-(C32, K4$\times$4, S2$\times$2),$IN$,LReLU \\

& $(\frac{h}{2},\frac{w}{2},32)$ & CONV-(C64, K4$\times$4, S2$\times$2),$IN$,LReLU \\

& $(\frac{h}{4},\frac{w}{4},64)$ & CONV-(C128, K4$\times$4, S2$\times$2),$IN$,LReLU \\

& $(\frac{h}{8},\frac{w}{8},128)$ & CONV-(C256, K4$\times$4, S2$\times$2),$IN$,LReLU \\

& $(\frac{h}{16},\frac{w}{16},256)$ & CONV-(C256, K4$\times$4, S2$\times$2),$IN$,LReLU \\

& $(\frac{h}{32},\frac{w}{32},128)$ & FC-(C256)\\
\hline
\hline
\multirow{7}*{Decoder} & (256) & FC-(C256 $\times \frac{h}{32} \times \frac{w}{32})$ \\

~ & $(\frac{h}{32},\frac{w}{32},256)$ &  DECONV-(C128,K4$\times$4,S2$\times$2),$IN$,LReLU \\

~&  $(\frac{h}{16},\frac{w}{16},128)$ &  DECONV-(C64,K4$\times$4,S2$\times$2 ),$IN$,LReLU \\

~&  $(\frac{h}{8},\frac{w}{8},64)$  &  DECONV-(C32,K4$\times$4,S2$\times$2),$IN$,LReLU \\

~&  $(\frac{h}{4},\frac{w}{4},32)$  &  DECONV-(C16,K4$\times$4,S2$\times$2),$IN$,LReLU \\

~&  $(\frac{h}{2},\frac{w}{2},16)$  &  DECONV-(C16,K4$\times$4,S2$\times$2),$IN$,LReLU  \\

~ & $(h,w,16)$ & CONV-(C3, k7$\times$7, S1$\times$1),Tanh \\
\hline
\end{tabular}
\end{center}
\caption{The architectures of Encoder and Decoder in generator $G_{x}$. The mask $M$ would be also as input of $G_{x}$.}
\label{tab:gx}
\end{table}

\begin{table}[h]
\begin{center}
\begin{tabular}{|lll|}
\hline
Module & Input Shape  & Layer Information \\
\hline\hline
\multirow{5}*{Encoder} & $(\frac{h}{2},\frac{w}{2},3)$ & CONV-(C16, K7$\times$7, S1$\times$1),$IN$,LReLU \\

& $(\frac{h}{2},\frac{w}{2},16)$ & CONV-(C32, K4$\times$4, S2$\times$2),$IN$,LReLU \\

& $(\frac{h}{4},\frac{w}{4},32)$ & CONV-(C64, K4$\times$4, S2$\times$2),$IN$,LReLU \\

& $(\frac{h}{8},\frac{w}{8},64)$ & CONV-(C128, K4$\times$4, S2$\times$2),$IN$,LReLU \\

& $(\frac{h}{16},\frac{w}{16},128)$ & FC-(C256) \\

\hline
\multirow{4}*{Decoder} & $(\frac{h}{16},\frac{w}{16},256)$ &  DECONV-(C128,K4$\times$4,S2$\times$2),$IN$,LReLU \\

~&  $(\frac{h}{8},\frac{w}{8},128)$ &  DECONV-(C64,K4$\times$4,S2$\times$2),$IN$,LReLU \\

~&  $(\frac{h}{4},\frac{w}{4},64)$  &  DECONV-(C32,K4$\times$4,S2$\times$2),$IN$,LReLU \\

~ & $(\frac{h}{2},\frac{w}{2},16)$ & CONV-(C3, k7$\times$7, S1$\times$1),Tanh \\
\hline
\end{tabular}
\end{center}
\caption{The architectures of the proposed self-supervised eye-flipping autoencoder $G_{pre}$. And it will be pretrained first.}
\label{tab:gpre}
\end{table}

\begin{figure}[h]
\section{CelebAGaze dataset.}
\begin{center}
\includegraphics[width=1.0\linewidth]{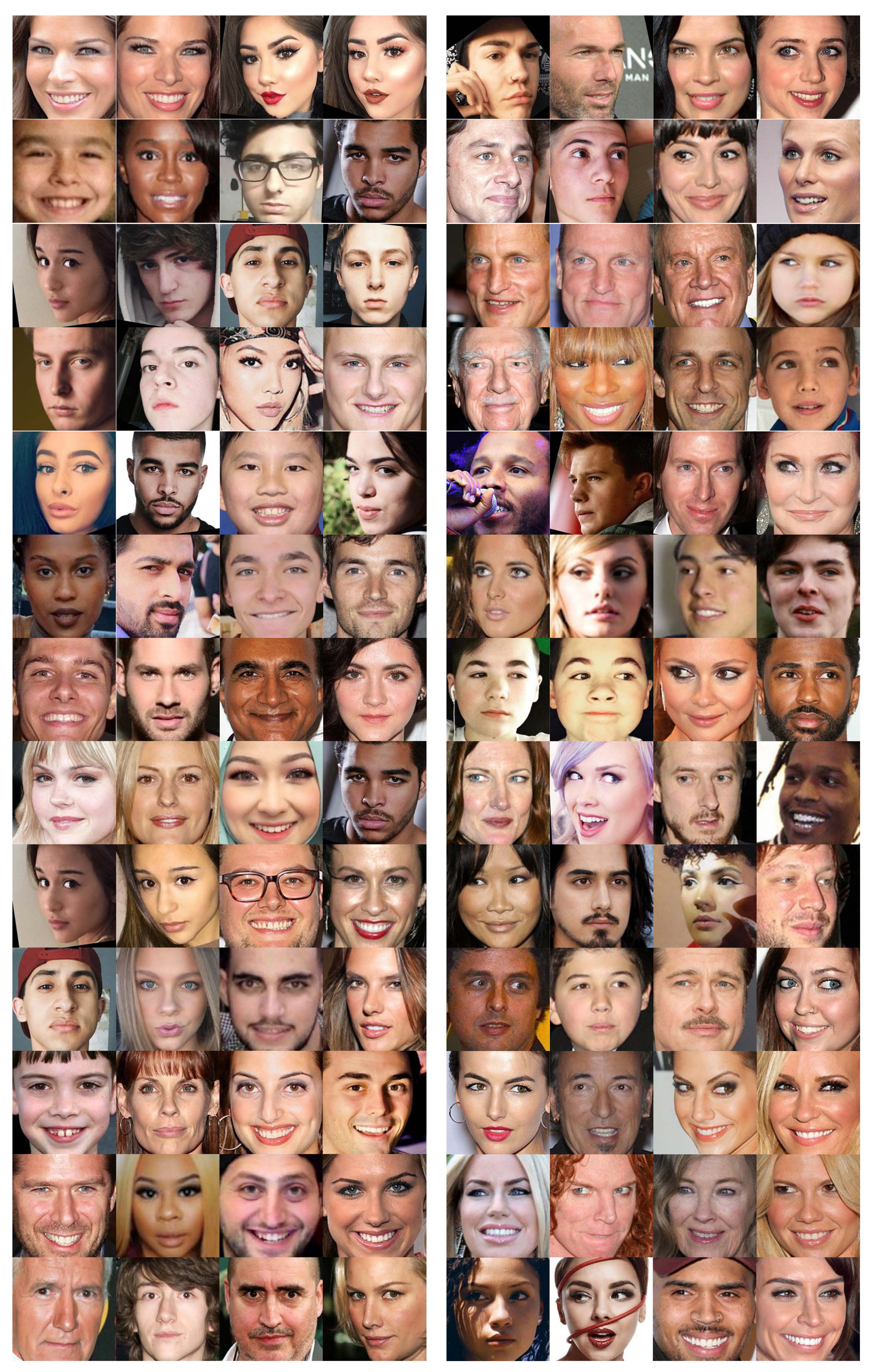}
\end{center}
\vspace{-0.5cm}
\caption{More examples from our CelebAGaze dataset. Left is from domain $X$ with staring at the cameras for gaze. Right is from domain $Y$ with staring at somewhere else for gaze.}
\vspace{-0.5cm}
\label{fig:newgaze}
\end{figure}

\end{document}